\documentclass[letterpaper, 10 pt, journal,twoside]{IEEEtran}

\usepackage{epsfig}
\let\labelindent\relax
\usepackage{graphicx}
\usepackage{bm}
\usepackage{amsmath}
\usepackage{amssymb}
\usepackage{enumitem}
\usepackage{booktabs}
\usepackage{amssymb}
\usepackage{enumitem}
\usepackage{booktabs}
\usepackage{xcolor}
\usepackage{multirow}
\usepackage{longtable}
\usepackage{adjustbox}
\usepackage{subfigure}
\usepackage{soul,color}
\usepackage{floatrow}
\usepackage{blindtext}
\usepackage{hyperref}
\usepackage{wrapfig,lipsum,booktabs}
\newfloatcommand{capbtabbox}{table}[][\FBwidth]
\newcommand{\subparagraph}{}
\usepackage[compact]{titlesec}
\usepackage{makecell}

\definecolor{Corr}{rgb}{1.0,0.0,0.0}
\definecolor{Typo}{rgb}{0.0,0.0,1.0}

\begin{document}

\title{Gated Recurrent Fusion to Learn Driving Behavior from Temporal Multimodal Data}
\author{Athma Narayanan$^{1}$, Avinash Siravuru$^{2}$, and Behzad Dariush$^{1}$%
\thanks{Manuscript received: September, 10, 2019; Revised December, 12, 2019; Accepted January, 9, 2020.}
\thanks{This paper was recommended for publication by Editor Eric Marchand upon evaluation of the Associate Editor and Reviewers' comments.
This work was supported by Honda Research Institute, USA.} 
\thanks{$^{1}$Athma Narayanan and Behzad Dariush are with Honda Research Institute, USA
{\tt\footnotesize \string{anarayanan, bdariush\string}@honda-ri.com}}%
\thanks{$^{2}$Avinash Siravuru is with Department of Mechanical Engineering, Carnegie Mellon University, USA
{\tt\footnotesize avinashs@cmu.edu}}%
\thanks{Digital Object Identifier (DOI): see top of this page.}
}
\markboth{IEEE Robotics and Automation Letters. Preprint Version. Accepted January, 2020}
{Narayanan \MakeLowercase{\textit{et al.}}: Gated Recurrent Fusion} 

\maketitle

\begin{abstract}
The Tactical Driver Behavior modeling problem requires an understanding of driver actions in complicated urban scenarios from rich multimodal signals including video, LiDAR and CAN signal data streams. However, the majority of deep learning research is focused either on learning the vehicle/environment state (sensor fusion) or the driver policy (from temporal data), but not both. Learning both tasks jointly offers the richest distillation of knowledge but presents challenges in the formulation and successful training. In this work, we propose promising first steps in this direction. Inspired by the gating mechanisms in Long Short-Term Memory units (LSTMs), we propose Gated Recurrent Fusion Units (GRFU) that learn fusion weighting and temporal weighting simultaneously. We demonstrate it's superior performance over multimodal and temporal baselines in supervised regression and classification tasks, all in the realm of autonomous navigation. On tactical driver behavior classification using Honda Driving Dataset (HDD), we report $10\%$  improvement in mean Average Precision (mAP) score, and similarly, for steering angle regression on TORCS dataset, we note a $20\%$ drop in Mean Squared Error (MSE) over the state-of-the-art.
\end{abstract}


\section{INTRODUCTION}

\IEEEPARstart{R}{ecently}, the domain of autonomous driving has emerged as one of the hotbeds for deep learning research, bolstered by strong industry support and availability of large real-world datasets (such as KITTI \cite{kitti}, Berkeley Driving Dataset \cite{bdd}, Honda Driving Dataset \cite{ramanishka2018toward}, Argoverse \cite{argoverse}) and physically/visually realistic simulators (\cite{wymann2000torcs, GymTORCS}, \cite{udacity}, \cite{carla}). Multi-sensor\footnote{In this paper, we use the terms \emph{sensor} and \emph{mode} interchangeably. The term \emph{sensor} is meaningful to interpret in the autonomous driving setting whereas  \emph{mode} is generally used in literature to indicate the various forms of state representation - this may come directly from sensors (image, speech signal) or after some meaningful post-processing (depth map, n-grams, etc.). } temporal data, provided by these datasets, offers more leverage to understand optimal driving actions. 

The current strategy for multimodal temporal data is to either pre-concatenate (concatenation followed by recurrent modules) \cite{ramanishka2018toward, lidar-video, girdhar2017actionvlad} or post-concatenate (parallel recurrent modules for each sensor followed by concatenation) \cite{ren2016look-aaai, jain2016recurrent, CorrRNN, morency-multiview, girdhar2017actionvlad}. While they both offer unique merits and challenges, in neither of the two choices, the multi-sensor data is \emph{fused explicitly}. Moreover, such design choices may lead to networks with much bigger parametric spaces resulting in training difficulties. Hence designing efficient architectures to exploit this rich source of data is still an open research problem. 
\begin{figure}[t]
\centering
\includegraphics[width=0.8\columnwidth]{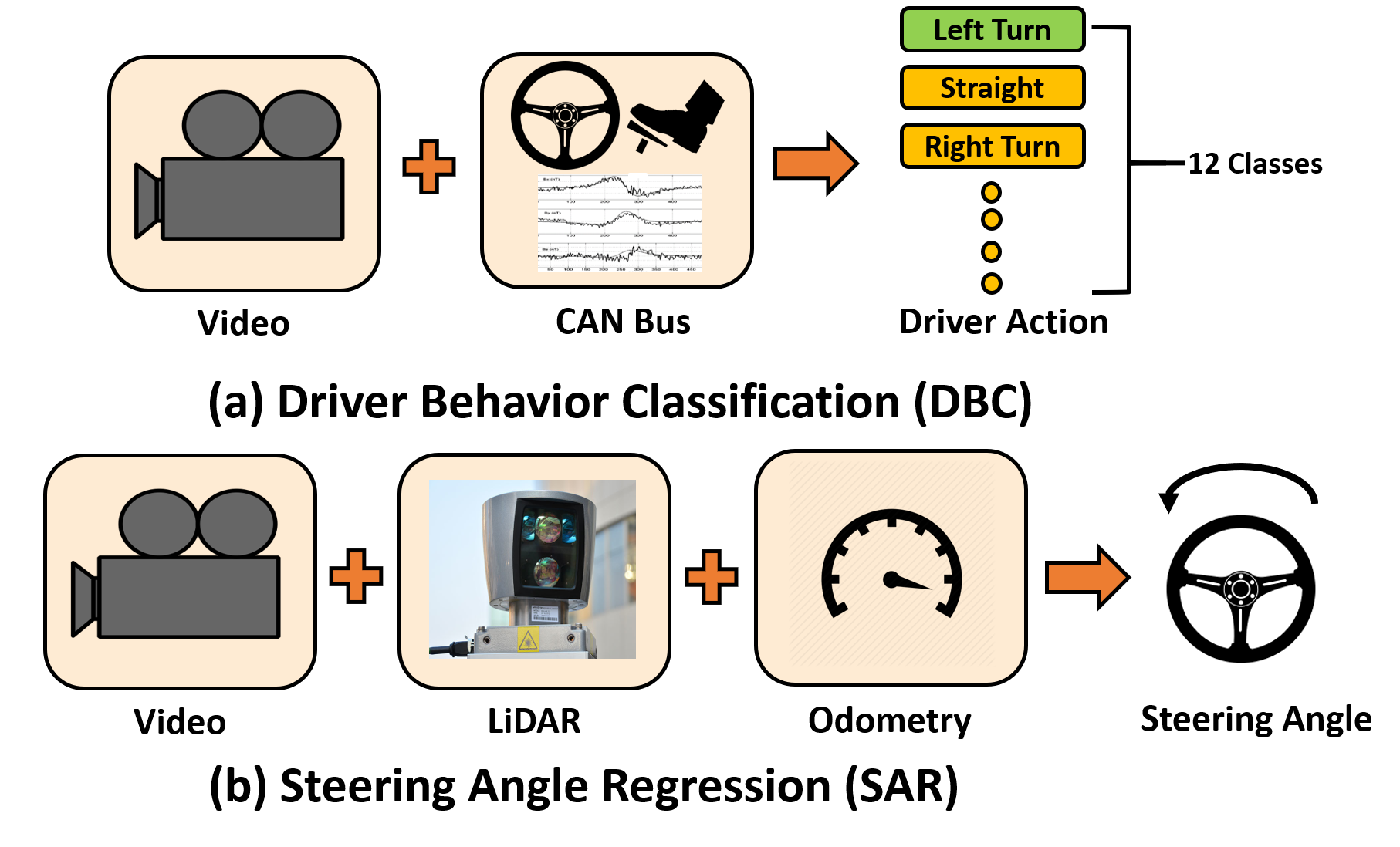}
\caption{In (a) \textbf{Driver Behavior Classification}, the objective is to correctly classify driver action based on given video and CAN data streams into one of the twelve ground-truth labels and, in (b) \textbf{Steering Angle Regression}, we predict the correct steering angle from video, LiDAR and odometry data streams. Both are used to understand and compare driving styles.}
\label{fig:into}
\end{figure}

Research on temporal fusion is popular in multimedia domains where either text or audio is combined with video \cite{CorrRNN, ren2016look-aaai}. However, in the autonomous driving community, while non-temporal fusion is popular (\cite{pomerleau1989alvinn, muller2006off,gao2018object}), temporal fusion has garnered much lesser attention. Compared to multimedia datasets, autonomous driving datasets present additional challenges, namely,  
\begin{itemize}
  \item There can be much more number of sensor inputs or multiple copies of each  sensor type (multi-camera \cite{nvidiacar} and multi-radar, etc.)
  \item The individual sensor data sizes could be disproportionate leading to undesirable biases towards a select few
  \item Intermittent data quality degradation or sensor failures could occur (e.g., motion blur and occlusion in image, LiDAR failures in snow, etc.)
\end{itemize}

This makes autonomous navigation a more general and challenging setting for developing temporal fusion models. Therefore, we first validate our models on autonomous driving related tasks. Given these complex inter-dependencies that emerge from learning on multimodal temporal data, it is essential to ensure that the models are interpretable to verify and correct for any over-fitting. Hence, in this work, fusion is formulated as the problem of finding the optimal linear interpolation between the sensors. The interpolation weights (also learned using \emph{gating} functions) can be interpreted as each sensor's percentage contribution to the fused state.



\textbf{Contributions of this work}: 
We introduce a novel recurrent neural network unit, called the Gated Recurrent Fusion Unit (GRFU) that can jointly learn fusion and target prediction from temporal data. This new formulation, designed to learn a linear interpolation of sensor encodings, offers superior performance two tasks (\emph{driver behavior classification} and \emph{steering angle regression}) on two challenging datasets (one \emph{real-world} and one \emph{simulated}, respectively) .
 By using global average pooling along the time dimension, we can explain the individual sensor’s contribution to the fused representation. Such interpretable fusion allows for higher-level intervention in the case of sensor failures. To the best of our knowledge, this is the first time this has been attempted in the autonomous navigation domain.

An overview of each task is shown in Fig. \ref{fig:into}. On the classification task, we report a \bm{$10\%$} improvement in the mAP score over the current state-of-the-art, and on the regression task, we note a \bm{$20\%$} drop in test error. 

The paper is structured as follows: In Section \ref{sec:litsurvey}, we review prior work in autonomous navigation and temporal fusion. In Section \ref{sec:methodology}, we formulate our Gated Recurrent Fusion Unit (GRFU) starting from a vanilla Long Short-Term Memory unit. Later, in Section \ref{sec:experiments}, we provide task-specific dataset details and analyze GRFU performance when compared to existing benchmarks. Finally, Section \ref{sec:conclusion}, summarizes the key ideas in the paper and lists future work avenues.

\section{Related Work}
\label{sec:litsurvey}
\subsection{Temporal Fusion in Machine Learning}
Learning using temporal multimodal data is a major sub-branch in deep learning research with applications that span video (\cite{ordonez2016deep,girdhar2017actionvlad}), audio (\cite{CorrRNN,ren2016look-aaai}) , text processing(\cite{morency-multiview,gallo2018image}) and robot navigation(\cite{kam1997sensor,sasiadek1999sensor}), to name a few. Various neural network architectures, like LSTM \cite{lstm}, GRU \cite{gru}, have been proposed to model and learn underlying data patterns temporally.  However, training complication becomes severe in a multimodal setting as complex correlations between the modalities and their history has to be disentangled.

Some of the major strategies for fusion include Kalman filtering~\cite{kam1997sensor}, and Fuzzy logic~\cite{sasiadek1999sensor}, just to name a few. These methods have shown great promise in robot navigation and are focused on combating sensor noise and sensor redundancy to do better state estimation. We, however, wish to propose an alternative strategy that can learn fusion using a more data-driven approach for driver behavior understanding. 

Most similar to our work in this are attention based fusion~\cite{hori2017attention} or LSTM modifications~\cite{ren2016look-aaai} used in multimedia data. Unlike explicit attention mechanisms that fuse non-temporal information outside the LSTM cell as in~\cite{hori2017attention}, we wish to focus on LSTM modifications within the LSTM cell unit. This we believe can model complex temporal correlations in a unified way (using state/information sharing between sensors).




\subsection{Learning in Autonomous Navigation}

Sensor fusion in autonomous navigation is a very active research topic~\cite{kam1997sensor}, \cite{sasiadek1999sensor}, \cite{li2013sensor}, \cite{gao2018object}, \cite{de2018fusion}. However,
temporal fusion has received much lesser attention even though they generate a lot of multimodal temporal data coming from a range of sensors like Camera, LiDAR (Light Detection and Ranging), wheel odometry, etc. Typical driving automation tasks of interest are learning driver behavior \cite{ramanishka2018toward} and intent \cite{jain2016recurrent}, motion forecasting \cite{luo2018fast, radwan2018multimodal}, object detection \cite{muller2006off}, learning affordances \cite{deepdriving}, action regression \cite{kendall2019learning, pomerleau1989alvinn}, semantic segmentation \cite{radwan2018vlocnet++, valada2018self} among others. 

Majority of research attacks these tasks in a non-temporal fashion, mainly using either RGB or RGB-D data \cite{pomerleau1989alvinn, muller2006off, radwan2018vlocnet++, deepdriving, nvidiacar}. Prior work on using fusion for autonomous navigation is either non-recurrent \cite{liu2017learning} in the reinforcement learning setting or recurrent unsupervised \cite{endtoendcars} in the motion forecasting setting. We present experiments showing that incorporating multimodal fusion, in a recurrent supervised setting is beneficial to model driver behavior.


\section{Methodology}
\label{sec:methodology}
In this Section, we describe the new temporal fusion architectures that we build over the standard LSTM model. We first review the LSTM model and simple fusion ideas in Section \ref{sec:ERF}. Next, in Section \ref{sec:ourmodels}, we introduce three new models that uses linear interpolation to find the optimal fused state to pass through the recurrent units\footnote{For easier visualization, all the model figures depicted in this Section are for two sensor case, but the equations are defined for an \emph{M} sensor case.}. 

\subsection{Preliminaries}
\label{sec:preliminaries}
Assume we are given a set of modalities $\bm{\Tilde{S}^1,\Tilde{S}^2,\Tilde{S}^3...\Tilde{S}^M}$ where \emph{M} is the  number of sensors, and the sensor signal for an arbitrary sensor, $i \in [1,M]$, is a time-series \({\bm{\Tilde{S}}^i = \big\{\Tilde{s}^{i}_{1},\Tilde{s}^{i}_{2}...\Tilde{s}^{i}_{\textbf{t}}...\Tilde{s}^{i}_{T} \big\}}\). The objective is to jointly learn the optimal temporal and modal composition to correctly predict the desired classification/regression target. 
Further, we make no additional assumptions like sensors having similar structure or dimensions, having similar forms of occlusions and noise ranges, or to be temporally correlated always. We do however pre-process all sensor inputs, $\Tilde{s}^i_t$ using appropriate encoders to bring them to the same dimension prior to temporal fusion (we call the processed sensor inputs as \emph{sensor encodings} and denote them as $s^i_\textbf{t}$). This is done for all proposed models and baselines for fair comparison.


The LSTM setup most commonly used in literature \cite{lstm, lstm-forgetgate} features three gated states (input $i_t$, forget $f_t$, output $o_t$) along with the hidden and candidate cell states $(h_t, c_t)$. Cell state represents memory while the hidden state is the output of the model at time $t$. The gated states control how much of the current and the past information need to be fused and transmitted to the next state in time. The two hidden states perform important functions namely:  slow-state $c_t$ that fights vanishing gradient problems, and a fast-state $h_t$ that allows the LSTM to make complex decisions over short periods of time. Each gated state performs a unique task of modulating the exposure and combination of the cell and hidden states. For a detailed overview of LSTM inner-workings and empirically evaluated importance of each gate, refer to \cite{bengio-empirical, zaremba-empirical}. 


\subsection{Early Recurrent Fusion (ERF)}
\label{sec:ERF}

\begin{figure}[h]
\centering
\includegraphics[width=\columnwidth,height=3.5cm,keepaspectratio=True]{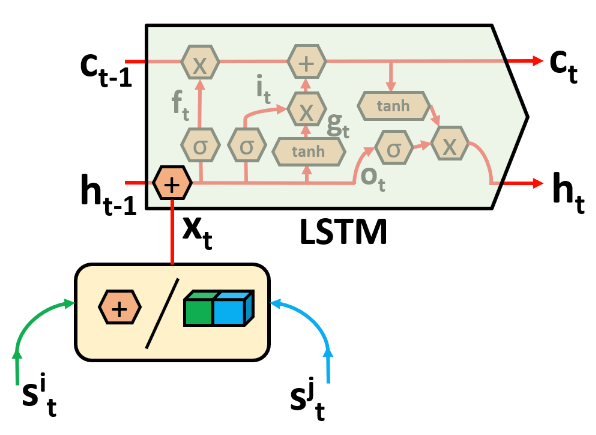}
\vspace*{-\baselineskip}
\caption{Early Fusion (Add/Concat) LSTM Unit}
\label{fig:earlyfusion-lstm-fig}
\end{figure}

A simple way to extend LSTMs to multimodal settings is by first summing or concatenating all the sensor encodings~\cite{ramanishka2018toward, lidar-video, girdhar2017actionvlad} and passing that as an input to the LSTM, ie.,  \({X = \big\{x_{1},x_{2},x_{3} \ldots x_{t} \big\}}\), where each $x_t$ = ($s^1 \oplus s^2 \ldots s^i \oplus s^j \ldots s^M$). From a temporal standpoint, one can view this as a type of early fusion. A simple ERF unit is shown in Fig.\ref{fig:earlyfusion-lstm-fig}.
\begin{align}
x_t &= \ldots s^i_{t} + s^j_{t} \ldots \ (\text{or}) \ \ldots s^i_{t} \oplus s^j_{t} \ldots,
\label{eqn:earlyfusion} \\ 
f_t &= \sigma(W_f \ast x_t + U_f  \ast h_{t-1}+ b_f), \nonumber \\
i_t &= \sigma(W_i \ast x_t + U_i \ast h_{t-1}+ b_i), \nonumber \\
o_t &= \sigma(W_o \ast x_t + U_o \ast h_{t-1}+ b_o), \nonumber \\
g_t &= \tanh(W_g \ast x_t + U_g \ast h_{t-1}+ b_g), \label{eqn:lstmgates}\\
c_t &= c_{t-1} \odot f_t + i_t \odot g_t, \nonumber \\
h_t &= o_t \odot \tanh(c_t).\label{eqn:lstmcellupdates}
\end{align}
\textbf{Remark:} Concatenation, while providing individual sensor inputs to the LSTM to extract useful information, bloats up the cell and hidden state size. On the other hand, summation reduces the cell size but naively combines all sensor encodings with equal emphasis. This may not be a good idea always, especially at time steps where one or more sensors provide noisy information to the fused state (for example, when a car is driving through a tunnel, camera information is unreliable). Temporal fusion architectures must be provided with sufficient tuning choice to learn how to fuse and use temporal data. This is necessary in driving datasets and both ERF models lack the explicit structures to learn them. Example scenarios where fusion needs to be \emph{dynamic} are, \\
1) \textit{\textbf{Occlusion in a sensor subset}:} While approaching an intersection a huge object in the form of a truck occludes the entire view in one of the image frames rendering image features unreliable. The model should rely on CAN data history to classify driver action correctly. \\
2) \textit{\textbf{Action dependency}:} Actions like lane branching are subtle steering actions. If the steering signal does not offer sufficient information, video features like lane markers and road curvature could supplement to avoid inter-class confusion. \\
3) \textit{\textbf{Loss of temporal correlation across sensors}:} As alluded to previously, when a car is going through a dark tunnel, optical flow for odometry maybe hard to obtain and might at best be weakly correlated to the data stream obtained from the CAN bus or LiDAR. Similarly, LiDAR gets really noisy and unreliable in snow, rain and grass \cite{lidarfailures}. 
\subsection{Proposed Temporal Fusion Models}
\label{sec:ourmodels}
We identify two important ways to mitigate the above mentioned issues, a) delay fusion and pass each sensor parallely through \emph{M} LSTM cells, allowing each sensor to individually decide how much of their respective histories to utilize with the current sensor input (we term this \emph{late recurrent sensor summation}), b) define gates for each sensor to determine the contribution of each sensor encoding to the fused cell and output states (we term this \emph{early gated recurrent fusion}). In the following Section, we first define both the modifications separately and finally define our main model which combines the two (this leads to the \emph{late gated recurrent fusion} model). Moreover, we use the \emph{late recurrent sensor summation} and \emph{early gated recurrent fusion} models also as baselines to evaluate the individual contributions (\emph{ablation study}) of the two modifications. 

\subsubsection{Late Recurrent Summation (LRS)}
\begin{figure}[b]
\centering
\includegraphics[width=\columnwidth,height=4.5cm,keepaspectratio=True]{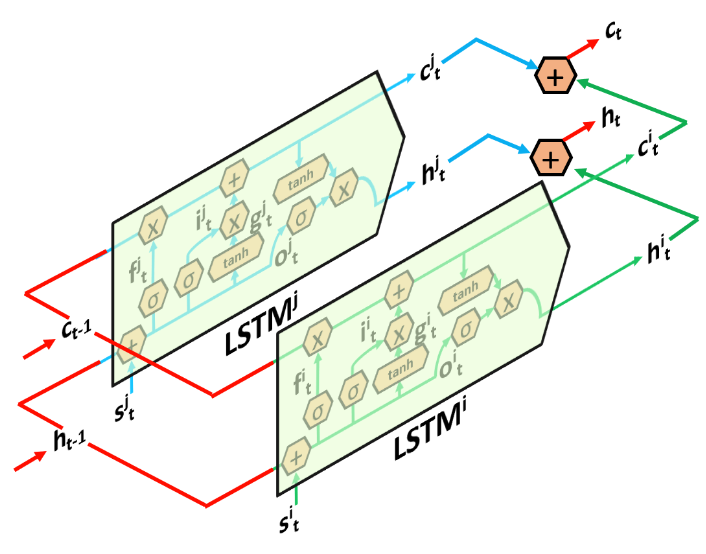}
\vspace*{-\baselineskip}
\caption{Late Recurrent Summation (LRS) LSTM Unit}
\label{fig:latefusion-lstm-sum}
\end{figure}

In this model, we use \emph{M} distinct LSTM units in total (one for each sensor). For each modality separate \emph{forget}, \emph{input} , \emph{output} and \emph{cell} states are computed. Model schematic with equations are shown in Fig.\ref{fig:latefusion-lstm-sum} and Eqns.(\ref{eqn:lrssgates})-(\ref{eqn:lrsssum}) respectively.
\label{sec:LRS}
\begin{align}
f_t^{i} &= \sigma(W_f^{i} \ast s_t^{i} + U_f^{i}  \ast \bm{h_{t-1}}+ b_f^{i}), \nonumber \\
i_t^{i} &= \sigma(W_i^{i} \ast s_t^{i} + U_i^{i} \ast \bm{h_{t-1}}+ b_i^{i}), \nonumber \\
o_t^{i} &= \sigma(W_o^{i} \ast s_t^{i} + U_o^{i} \ast \bm{h_{t-1}}+ b_o^{i}), \nonumber \\
g_t^{i} &= \tanh(W_g^{i} \ast s_t^{i} + U_g^{i} \ast \bm{h_{t-1}}+ b_g^{i}),
\label{eqn:lrssgates}\\
c_t^{i} &= \bm{c_{t-1}} \odot f_t^{i} + i_t^{i} \odot g_t^{i}, \nonumber \\
h_t^{i} &= o_t^{i} \odot \tanh(c_t^{i}),\label{eqn:lrssupdates} \\
\bm{c_t}  &= \sum_{i=1}^{M} c^{i}_t , \quad \bm{h_t} = \sum_{i=1}^{M} h^{i}_t.
\label{eqn:lrsssum} 
\end{align}

The weights, $W_*$, $U_*$, and biases, $b_*$ , that transform the input space for each gate are unique for each modality but are shared across time. As summarized in the previous Section, each LSTM unit receives information from the states of the past time step ($c^i_{t-1}$, $h^i_{t-1}$) and the input from the current time step, $s^i_t$. Now, instead of having separate states of each LSTM unit of a sensor, all the copies receive the same states ($c_{t-1}$, $h_{t-1}$) obtained from the previous time-step. Through this modelling choice we can propagate fused representations temporally. In contrast, in \cite{ren2016look-aaai}, the weights are shared between modalities but not states. By sharing the past cell state ($c_{t-1}$) across all sensors, the model can individually decide whether to retain or discard memory for each modality. Finally, all the hidden ($h_t^{i}$) and cell ($c_t^{i}$) states are added to produce a combined representation $h_t$ and $c_t$ that is sent to the next time step (hence the prefix \emph{late} to indicate late fusion in the model name). 

\subsubsection{Early Gated Recurrent Fusion (EGRF)}
\label{sec:EGRF}
\begin{figure}[t]
\centering
\includegraphics[width=\columnwidth,height=5cm,keepaspectratio=True]{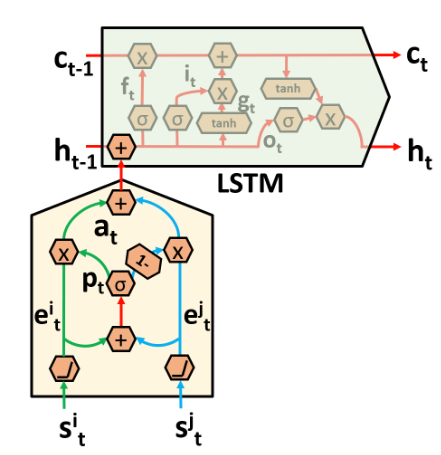}
\vspace*{-\baselineskip}
\caption{Early Gated Recurrent Fusion (EGRF) LSTM Unit}
\label{fig:earlygatedfusion-lstm}
\end{figure}

Late fusion offers the model some flexibility to separately control the memories of individual sensors, but even here summation at the end fuses all sensors assuming equal importance. However, we wish to also learn from the data the extent of each sensor's contribution to the final fused states. Inspired by the gating mechanisms used in the LSTM \cite{lstm,lstm-forgetgate} and GRU \cite{gru}, we propose a similar exposure control in the sensor fusion module as well. For \emph{M} sensors, we define \emph{M-1} gates ($p^*$) that control the exposure of the sensor encoding, $s^i_t$, in the final state $a_t$. Similar to \cite{gru}, we define the gating for the last sensor as $1 - \sum^{M-1}_i p^i$. This makes the joint representation a linear interpolation of individual sensor encodings. The model schematic and equations are shown in Fig.\ref{fig:earlygatedfusion-lstm} and Eqns.(\ref{eqn:fusion-embedding})-(\ref{eqn:fusion-updates}) respectively. Firstly, the sensors embeddings are converted to the same dimension using a non-linear operation as in Eqn.(\ref{eqn:fusion-embedding}). Then $M-1$ gates  are computed as shown in  Eqn.(\ref{eqn:fusion-gating}). As shown in Eqn.(\ref{eqn:fusion-sum}), the final fusion is performed where each gate is multiplied to the corresponding sensor encoding and summed to form the fused state $a_t$. Temporal Modelling is performed with $a_t$ as the input as shown in Eqns.(\ref{eqn:fusion-gates})-(\ref{eqn:fusion-updates}).
\begin{align}
e_t^{i} &= \text{relu}(W_e^{i} \ast s_t^{i}), \label{eqn:fusion-embedding}\\
p_t^{k} &= \sigma( \sum_{i=1}^{M} W_p^{i} \ast e_t^{i}), \forall k \in [1,M-1],\label{eqn:fusion-gating} \\
\bm{a_t} &= (\sum_{k=1}^{M-1} p_t^k \odot e_t^k) + (1-\sum_{k=1}^{M-1}p_t^k) \odot e_t^k, \label{eqn:fusion-sum}\\
f_t &= \sigma(W_f \ast \bm{a_t} + U_f  \ast h_{t-1}+ b_f), \nonumber \\
i_t &= \sigma(W_i \ast \bm{a_t} + U_i \ast h_{t-1}+ b_i), \nonumber \\
o_t &= \sigma(W_o \ast \bm{a_t} + U_o \ast h_{t-1}+ b_o), \nonumber \\
g_t &= \tanh(W_g \ast \bm{a_t} + U_g \ast h_{t-1}+ b_g), \label{eqn:fusion-gates}\\
c_t &= c_{t-1} \odot f_t + i_t \odot g_t, \nonumber \\
h_t &= o_t \odot \tanh(c_t).
\label{eqn:fusion-updates}
\end{align}
The gating functions are valuable to draw insights and explain the nature of fusion occurring within the model. Once learnt, the user can interpret the gating values as contributions of each sensor and verify if they match human insight for some arbitrary sample in the dataset. This explainability feature is crucial for scenarios involving safety-critical tasks. 
\subsubsection{Late Gated Recurrent State Fusion (LGRF)}
\label{sec:LGRF}
Finally, we describe our Late Gated Recurrent Fusion model, which combines the best aspects of both late recurrent fusion (independent control of memory for each sensor) and gated recurrent fusion (learning how to fuse) in order to improve learning performance of temporal fusion models.
\begin{align}
e_t^{i} &= \text{relu}(W_e^{i} \ast s_t^{i}). \label{eqn:multisesnor-fusion-embedding}\\
p_t^{k} &= \sigma( \sum_{i=1}^{M} W_p^{i} \ast e_t^{i}). \quad \forall k \in [1,M-1].\label{eqn:multisensor-fusion-gating} \\
\bm{a_t^{i}}  &=
\begin{cases}
p_t^i \odot e_t^i & \text{if}\quad i \in [1,M-1],\\
(1-\sum_{k=1}^{M-1}p_t^k) \odot e_t^i & \text{if}\quad  i=M\\
\end{cases} \nonumber \\
f_t^{i} &= \sigma(W_f^{i} \ast \bm{a_t^{i}} + U_f^{i}  \ast \bm{h_{t-1}}+ b_f^{i}), \nonumber \\
i_t^{i} &= \sigma(W_i^{i} \ast \bm{a_t^{i}} + U_i^{i} \ast \bm{h_{t-1}}+ b_i^{i}), \nonumber \\
o_t^{i} &= \sigma(W_o^{i} \ast \bm{a_t^{i}} + U_o^{i} \ast \bm{h_{t-1}}+ b_o^{i}), \nonumber \\
g_t^{i} &= \tanh(W_g^{i} \ast \bm{a_t^{i}} + U_g^{i} \ast \bm{h_{t-1}}+ b_g^{i}), \label{eqn:multisensor-fusion-gates}\\
c_t^{i} &= \bm{c_{t-1}} \odot f_t^{i} + i_t^{i} \odot g_t^{i}, \nonumber \\
h_t^{i} &= o_t^{i} \odot \tanh(c_t^{i}),
\label{eqn:multisensor-fusion-updates}\\
\bm{c_t}  &= \sum_{i=1}^{M} c^{i}_t , \bm{h_t} = \sum_{i=1}^{M} h^{i}_t.
\label{eqn:multisensor-fusion-sum}
\end{align}
\begin{figure}[t]
\centering
\includegraphics[width=\columnwidth,height=5.3cm,keepaspectratio=True]{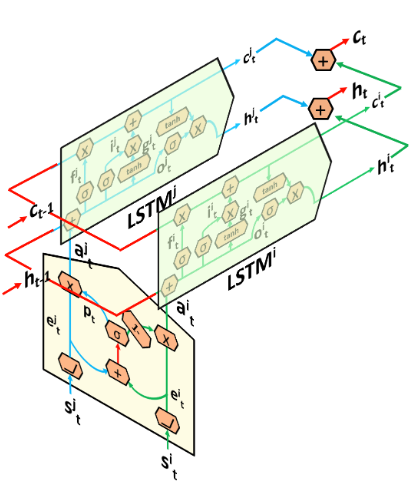}
\vspace*{-\baselineskip}
\caption{Late Gated Recurrent Fusion (LGRF) LSTM Unit}
\label{fig:lategatedfusion-lstm}
\end{figure}
The model schematic is shown in Fig.\ref{fig:lategatedfusion-lstm}. Similar to the early gated recurrent fusion model, we compute fusion gates $p_t^*$ as a function of all the sensor encodings $e_t^*$, but instead of doing the linear interpolation of all sensor inputs to get a joint input state, $a_t$, we use the gates to control the exposure of each encoding that is passed into sensor specific LSTM cells. The final joint cell and hidden states are computed by summing all the final cell and hidden state outputs.
Having described the new temporal fusion designs, in the next Section, we test the models on two challenging autonomous driving datasets.

\section{Experiments}
\label{sec:experiments}
\subsection{Tactical Driver Behavior Classification}
\subsubsection{HDD Dataset}
Recently, HDD~\cite{ramanishka2018toward} was proposed to stimulate research on learning driver behavior in interactive situations. The dataset includes a 104-hour synchronized multi-sensor naturalistic driving data. We focus our attention on the \textbf{goal-oriented} driver behavior classification task which involves temporally classifying the multimodal data involving video stream and CAN signal data into driver actions. The 104-video-hour data corresponds to 137 sessions. Each frame contains one label from the twelve behavior classes such as \textit{left turn, right turn, intersection passing, lane change, etc.}

We follow the prior work~\cite{ramanishka2018toward} and obtain our training (100 driving sessions) and testing splits (37 driving sessions).  CAN signal includes: car speed, accelerator and braking pedal positions, yaw rate, steering wheel angle, and the rotation speed of the steering wheel,turn signals (eight dimensional stream). The images are of dimension $ 720 \times 1280 \times 3 $.
The image representation is extracted from conv$2d_7b_{1x1}$ layer of InceptionResnet-V2 \cite{szegedy2017inception} pre-trained on ImageNet \cite{deng2009imagenet}. The features are convolved with a $1\times1$ convolution to reduce the dimension from $8\times8\times1536$ to $8\times8\times20$ and  flattened to $1\times1280$. Raw sensor signals are passed through a fully-connected layer to  transform $1\times8$ size signal to obtain a feature vector of size $1\times20$. 

\subsubsection{Results}
In this task, the input is untrimmed, egocentric sequences of video and CAN signals. The output is the tactical driver behavior label of each frame. We follow the evaluation protocol as in~\cite{NakamuraCVPR2017,ramanishka2018toward,shou2018online} to compute  mean Average Precision (mAP) over all classes. We use the Adam optimizer~\cite{kingma2014adam} to learn the network parameters with the sequence length set to 90 video frames. To fairly compare with the baseline methods~\cite{ramanishka2018toward}, we use the same batch size set to 40. The training is performed using truncated back-propagation through time. The training process is terminated after 50 epochs, with a fixed learning rate $5\times10^{-4}$. The results for the same test split as used in ~\cite{ramanishka2018toward} are showcased in Table.\ref{table:results}.\\ 
{\textbf{Non-Fusion Architecture.}} We first perform experiments only on the  \textit{CAN} signal and \textit{Img} (Image) sensors separately. The embeddings are directly sent to a standard LSTM with hidden size of dimension $2000$. The output $h_t$ is directly fed into a fully connected layer then squashes the dimension to $12$ classes including \textit{background class.} The CAN signal outperforms in certain classes such as \textit{left turn, right turn, U-turn} while Image performs better in classes such as \textit{lane change, lane branch, intersection passing and crosswalk passing}. TCN \cite{bai2018empirical} performs slightly better than LSTM as showcased in Table.\ref{table:results}. A successful sensor fusion should outperform these results benefiting from each sensor separately.\\
{\textbf{{Early Fusion LSTM}.}} As baseline architectures we use the early sensor fusion where sensor embeddings are either concatenated (\textit{Early-Concat}) or element wise summed (\textit{Early-Add}) as explained in Section \ref{sec:ERF}. \textit{ Early-Concat} is similar to the technique used in \cite{ramanishka2018toward}. In the early fusion stage the \textit{Early-Concat} outperforms \textit{Early-Add} (mAP of 32.66 vs 29.88) as the LSTM has access to individual sensor information, and can choose to discard noisy sensor readings. However each sensor has different ranges and are normalized individually. For example, steering angle is between +360$^{\circ}$ to -360$^{\circ}$ and image has pixel values between 0 to 255. Features added from a normalized 0 pixel value and 10$^{\circ}$ steering input can result in the same output when the values are interchanged. Several such combination of values can be created. Hence adding could corrupt the fused encoding resulting in the LSTM operating on a corrupted feature space.\\
{\textbf{Late Fusion LSTM}.}
Here we have two separate LSTM cells that do not share any weights or hidden states between the modalities. Concatenation or summation happens after the LSTM cell. More precisely $h_t^{Img} \oplus h_t^{CAN}$ is sent to a single fully connected layer for classification. The fully connected layer operates on a $2000 \times 2$ dimension vector in the case of \textit{Late-Concat} or $2000$ dimension vector in the case of \textit{Late-Add} respectively. Interestingly \textit{Late-Add} (which is essentially LRS without cell state sharing) outperforms all other types of baseline fusion as the addition of cell states allows the model to focus more on the temporal aspects of each sensor.\\
{\textbf{Look, Listen and Learn \cite{ren2016look-aaai}}.} The most similar baseline to our \textit{LRS} model described in Section \ref{sec:LRS} is the Look, Listen and Learn architecture presented in \cite{ren2016look-aaai}. We re-implement the architecture in Pytorch for the HDD dataset. We add auxiliary losses to both the modalities and sum in the predicted output results with weight sharing. This results in a huge improvement over baseline model over every class. We call this model \textit{Look Listen} in Table.\ref{table:results}\\
{\textbf{EGRF, LRS, LGRF.}} We replace the standard LSTM with each of our fusion modules explained in Section \ref{sec:EGRF} (\textit{EGRF}) , Section \ref{sec:LRS} (\textit{LRS}) , Section \ref{sec:LGRF} (\textit{LGRF}). Each of our hypothesised fusion architectures outperform the state of the art on almost all of the classes. Our \textit{EGRF} and \textit{LRS} models increase the mAP by $ 7 \%$ over the standard fusion LSTMs while benefiting distinct class labels as shown in Table.\ref{table:results}. Finally, we hypothesize that our combined model \textit{LGRF} attempts to combine the benefits of both \textit{LRS} and \textit{EGRF} and therby increasing the mAP by $ 10 \%$. The main driver for the performance boost is the added flexibility in learning afforded by the gating functions which allow the network to modulate the fusion process at each time step and best optimize the data being input from individual sensors. 

\begin{figure}[hb!]
\centering
\includegraphics[width=\columnwidth,height=10cm]{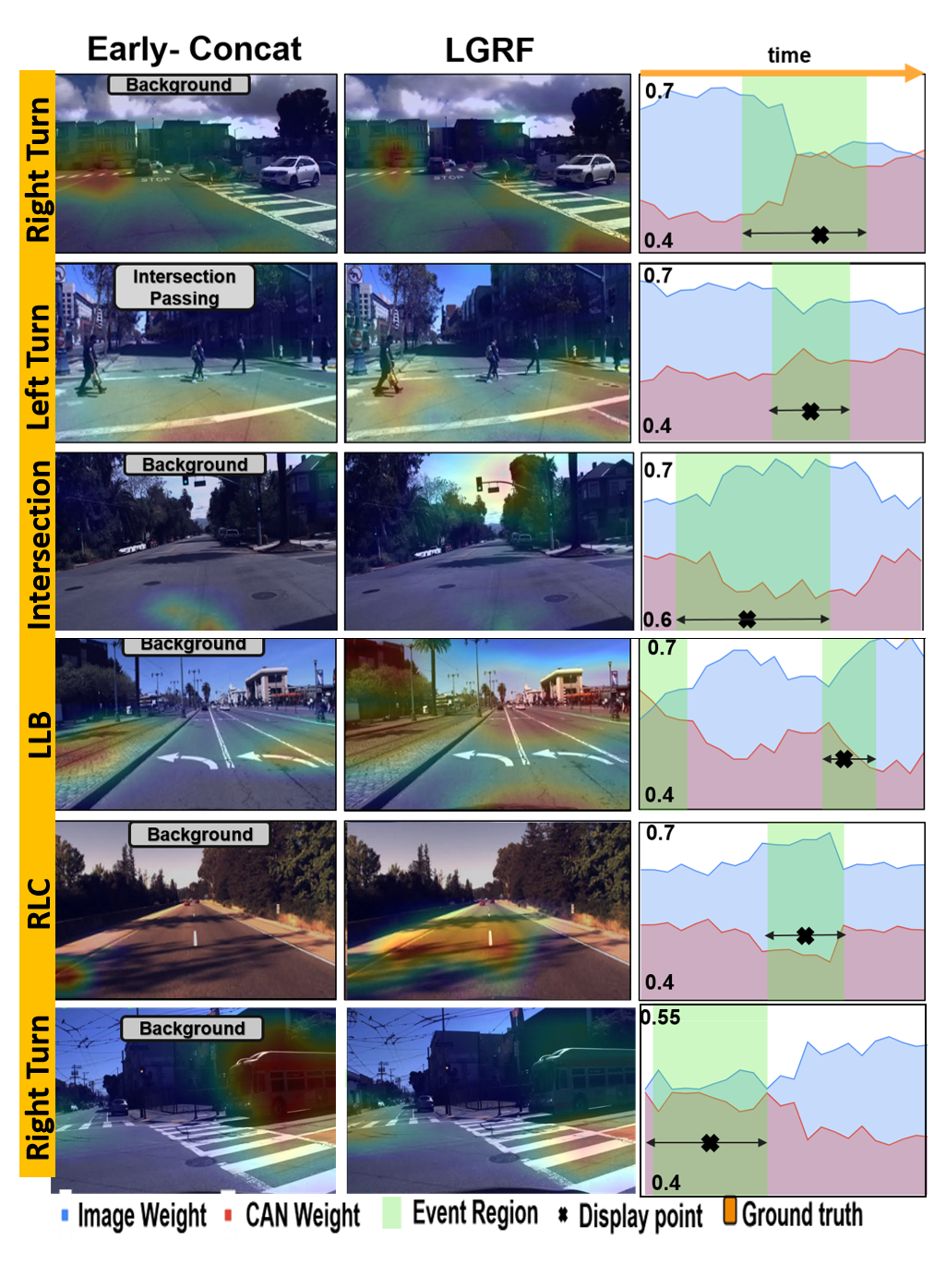}
\caption{Sensor Attention visualized for different actions where baseline models fail, and our model succeeds. In most of the actions we can see a shift in the attention to more meaningful parts of the image by LGRF model. For Right Lane Change (RLC) focus is on lane marker, Left lane branch (LLB) on branch arrow, Intersection on traffic lights and road extension. Moreover, we can also determine the weighting scheme the model used to predict the correct results using global average pooling on the fusion gates as shown in the last column. (Shown in Eq.\ref{eqn:fusion-gating}).}
\vspace*{-\baselineskip}
\label{fig:attention}
\end{figure}
\begin{table*}[ht]
\centering
\setlength{\tabcolsep}{4pt}
\setlength\arrayrulewidth{0.2pt}
\resizebox{\textwidth}{!}{
\begin{tabular}{ccccccccccccccc}
\toprule
Models & Fusion & Inputs &
\makecell{intersection \\ passing} &
\makecell{left \\ turn} &
\makecell{right \\ turn} & 
\makecell{left \\ lane \\ change} &
\makecell{right \\ lane \\ change} &
\makecell{left \\ lane \\ branch } &
\makecell{right \\ lane \\ branch} &
\makecell{crosswalk \\ passing   } &
\makecell{railroad \\ passing    } &
\makecell{merge                  } &
\makecell{u-turn                 } &
mAP   \\
\midrule
\multirow{7}{*}{Non-Fusion} 
& - & CAN     & 36.41 & 66.25 & 74.27 & 26.17 & 13.39 &  8.03 &  0.20 &  0.30 &  0.07 &  3.59 & 33.57 & 23.84 \\
& - & Img     & 65.74 & 57.79 & 54.43 & 27.84 & 26.11 & 25.76 &  1.77 & 16.08 &  2.56 &  4.86 & 13.65 & 26.96 \\
& Early-Concat \cite{ramanishka2018toward} & CAN+Img & 74.93 & 71.68 & 76.80 & 32.68 & 37.02 & 29.78 &  5.51 &  9.91 & 3.62 &  3.44 & 13.90 & 32.66 \\
& Early-Add & CAN+Img & 77.08 & 75.90 & 76.40 & 23.38 & 19.14 & 16.62 & 2.51 & 10.19 & 2.67 & 8.90 & 15.91 & 29.88 \\
& Late-Concat  & CAN+Img & 70.04 & 77.43 & 75.82 & 29.98 & 13.37 & 21.89 & 3.08 & 4.70 & 1.98 & 2.88 & 30.06 & 30.11 \\
& Late-Add  & CAN+Img & 79.03 & 76.84 & 76.64 & 28.61 & 21.72 & 20.40 & 4.29 & 12.15 & 2.50 & 7.66 & 31.30 & 32.83 \\
& TCN \cite{bai2018empirical}  & CAN+Img & 79.98 & 71.91 & 73.80 & 28.14 & 23.60 & 30.69 & 4.63 & 8.46 & 2.99 & 7.96 & 32.11 & 33.16 \\
\midrule
\multirow{4}{*}{\makecell{Fusion \\ LSTM} } 
& {Look,Listen \& Learn \cite{ren2016look-aaai} }  & CAN+Img & 81.11 & 78.46 & 79.01 & 43.20 & 25.29 & 30.17 & \textbf{7.79} & 13.94 & 3.56 & 8.92 & 33.39 & 36.80 \\
& \textbf{EGRF(ours)}   & CAN+Img & 84.29 & 82.36 & 80.21 & 48.26 & 29.67 & \textbf{40.39} & 5.69 & 18.23 & 3.06 & 9.70 & \textbf{37.93} & 39.98 \\
& \textbf{LRS(ours)} & CAN+Img & 82.40 & 77.04 & 76.49 & 51.77 & 35.14 & 32.85 & 6.66 & 18.29 & \textbf{3.82} & \textbf{11.00} & 36.74 & 39.29 \\
&\textbf{ LGRF(ours) }& CAN+Img & \textbf{86.83} & \textbf{85.39} & \textbf{82.95} & \textbf{57.76} & \textbf{37.79} & 37.42 & 3.85 & \textbf{20.38} & 3.35 & 10.60 & 37.07 & \textbf{42.13} \\
\bottomrule
\end{tabular}
}
\vspace*{-\baselineskip}
\caption{mAP of Driver behavior classification on HDD dataset }
\label{table:results}
\end{table*}

\subsubsection{Discussion}
One of the limitations of most sensor fusion architectures is the inability to provide visual explanations for the decision-making process. For example, when in the case of a noisy sensor signal, the model needs to adapt to another sensor and gate the noise. LGRF model is uniquely positioned to give class specific reasoning for the sensor weighting. For this, we apply global average pooling on the pre-gate layer $p_k^{si}$ along the sensor dimension and display its value. For example, a value of $0.7$ for sensor one means that sensor one had a higher weighting than the $0.3$ for sensor two.  We additionally visualize the class activation maps~\cite{zhou2016learning} to show the localization ability of our models by using Grad-CAM~\cite{gradcam} on the last convolution layer of the image input.

We get explainable results that validate our assumptions about which sensor is important for which action, as showcased in Fig.~\ref{fig:attention}. The heat map falls on image locations such as lane markers for lane actions, road extensions for turns or intersection passing. Turns have higher CAN weighting as they capture the motion better. An interesting observation is the truck occluding the view in the last example. Our model not only improves the attention region by localizing to the cross-walk but also shows equal weighting for both images and CAN signals, thereby correctly classifying the action.

\subsubsection{Failure cases} 
\begin{figure}[h]
\begin{center}
  \includegraphics[width=\textwidth,height=4.7cm]{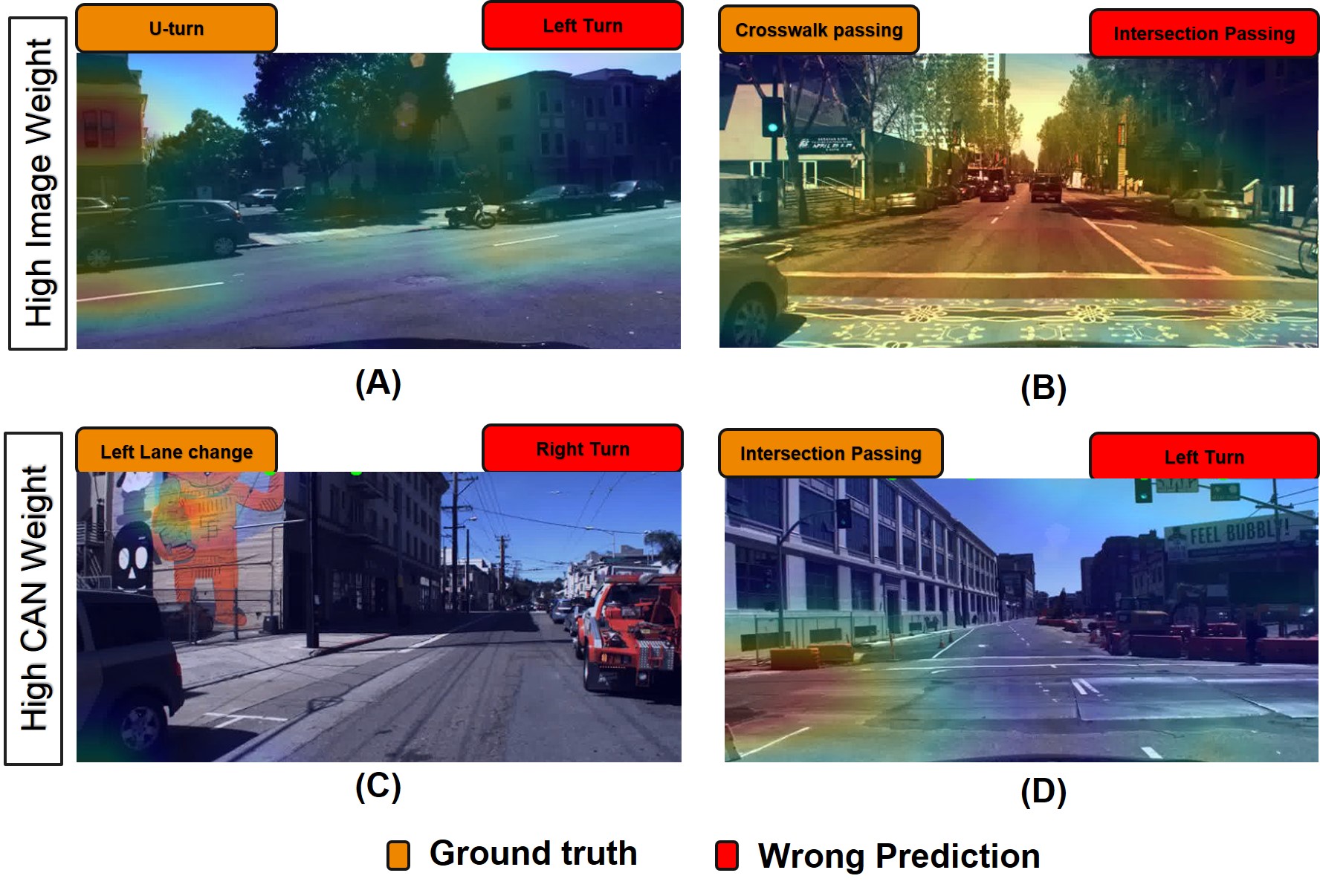}
\end{center}
\caption{Failure modes of our models. \textit{High Image weight:} (A) Inter class confusion between Left turn and U-Turn due to improper long term CAN signal weighting. (B) Crosswalk and Intersection confusion is high in spite of expected higher Img weight. \textit{High CAN weight:} (C) The Attention appears on graffiti. Higher CAN weight does not resolve issue. (D) Attention falls on plausible road extension but is misclassified due to higher CAN weight.}
\vspace*{-\baselineskip}
\label{fig:failure modes}
\end{figure}

We focus on the driver behavior classification task and visualize the attention maps on the image sensor to better understand the failure cases. We conjecture that the majority of failure cases occur due to: 1) Wrong sensor being weighted higher, 2) Fusion model's inability to resolve inter class confusion. Such cases, as shown in Fig.\ref{fig:failure modes} , have to be investigated further to resolve unexplained attention regions.

\subsection{Steering Action Regression}
To show the generality of our methodology we test our models on steering angle regression as well. Given a set of sensor signals the task is to determine the appropriate steering control action to successfully drive in a race track. One method of addressing this problem is to perform end to end regression. A better temporal fusion could provide richer features to deal with the challenging task of understanding vehicle dynamics just by observing sensors. We also showcase an extension of our work to a three sensor setting.


\subsubsection{TORCS Dataset}
TORCS driving simulator is capable of simulating physically realistic vehicle dynamics as well as multiple sensing modalities to build sophisticated AI agents that can complete race tracks. The following sensing modalities for our state description include : (1) odometry (\emph{SpeedX, SpeedY, SpeedZ}) substituting CAN signals (2) laser scans consisting of 19 LiDAR points. (3) color images capturing the ego-view.
\begin{table*}[ht]
\centering
\setlength{\tabcolsep}{4pt}
\setlength\arrayrulewidth{0.2pt}
\begin{tabular}{c|cccccccccc}
\toprule
Model & Image & LiDAR  & Odometry & Early-Concat\cite{ramanishka2018toward} & TCN\cite{bai2018empirical}  & Late-Add & Look Listen\cite{ren2016look-aaai} & \textbf{EGRF} & \textbf{LRS} & \textbf{LGRF}\\
\midrule
MSE & 1.3973 & 1.5610 & 2.10 & 0.997 & 0.71 & 1.1051 & 0.810 & 0.7087 & 0.635 & \textbf{0.619} \\
\bottomrule
\end{tabular}
\vspace*{-\baselineskip}
\caption{MSE of steering angle regression on TORCS dataset}
\label{table:torcsresults}
\end{table*}
We collect $1000$ time steps  from 32 different tracks that vary in the form of complicated loops to different road conditions. To collect the steering action ground truths we use the standard PID controller that successfully completes navigating one lap on each track without veering off the road. Out of the 32 tracks, we divide the training-test into $20$-$12$ track split. We perform 5 fold cross validation on training data ( 80 - 20 split) and display the best model results for the test split in Table \ref{table:torcsresults}. For the encoding we employ multiple convolution layers of kernel size $3\times3$. More specifically, the following layers are used: \textit{layer1,layer2} has 32 filters, followed by max pooling with kernel size $2\times2$, followed by \textit{layer3,layer4} with 64 filters each. Finally an additional max pooling layer with $2\times2$ to downsample the feature space to a feature size of $1\times1638$. The $1\times19$ velodyne points and $1\times3$ odometry signals are embedded using separate linear layers to  $1\times30$ and  $1\times20$ respectively. The flattened feature embeddings are used as inputs to our model. The output of the LSTM is sent to a linear layer followed by $\tanh$ activation. 
\subsubsection{Results and Discussion}
In this task the input is a batch of images, odometry and LiDAR points with a time history of four time steps. We train the model to directly regress the steering action. We compute the average Mean squared error (MSE) between the predicted action and ground truth action value over all the data in test set. After ablation study we found that history above four time steps does not provide significant information for temporal modeling. We set the batch size to $128$ for all experiments and use Adam optimizer to train the weights. For the final activation we choose $\tanh$ activation function to squish the last linear layer output to a range $(-1,1)$. We perform a grid search on the learning rate from $1e-2$ to $1e-5$. Overall $5e-3$ performs best. The comparison of the overall MSE for the test set is shown in Table~\ref{table:torcsresults}.\\
\textbf{Baselines.} We extend our previous baselines from HDD experiments to a three sensor setting and report the results in Table \ref{table:torcsresults}. A similar trend in results is obtained with \cite{ren2016look-aaai} outperforming other baselines with the lowest error of $0.810$.  Most of the models are not able to handle the huge disparity in sensor embedding dimensionality with image embedding size of $1\times 1638$ overshadowing others. \\ 
\textbf{EGRF, LRS, LGRF.}
We extend our models to a three sensor setting. This involves 1) modifying Eqn.~\ref{eqn:lrssgates} in LRS to support 12 gates (4 for each modality) 2) compute two pre-gates for EGRF as in Eq.\ref{eqn:fusion-gating} for images ($p_t^{img}$), LiDAR  ($p_t^{lidar}$) and odometry ($1-p_t^{img} \times p_t^{lidar}$) 3) Combine both for LGRF. Our models outperform all other baselines. LGRF gives an overall best performance with an additional +20\% drop over the state-of-the-art. An interesting note is the huge variation in error between EGRF and LRS. We suspect that this may be due to the highly correlated sensors in a simulated setting as opposed to the real world setting in HDD dataset. Hence the benifit of early noise rejection from EGRF does not play as important a role as learning to fuse the best aspects of each sensor as in LRS. 

\section{Conclusion}
\label{sec:conclusion}
In this work we presented a novel temporal fusion architecture that we termed \emph{Gated Recurrent Fusion Unit} to learn from large-scale multi-sensory temporal data. Gating functions modulate the exposure of individual sensor data at each time step to determine optimal fusion strategy. GRFU  eliminates the need to design and train separate network blocks for pre-processing sensor data, learning intermediate representations, or driver behavior modeling assuming full state information. As GRFU is end-to-end differentiable, all these building blocks can be learned together. 

For future work we plan to evaluate GRFU's effectiveness on other challenging temporal multimodal settings not limited to autonomous driving domain. Moreover additional extensions using TCN backbones may be considered to determine the best temporal abstraction for fusion. Additionally, for situations where the data is not synchronized, or, actions depend on tracking of particular objects in the scene, we could add  attention  mechanisms \cite{morency-multiview} to GRFU to provide more context.


\bibliographystyle{IEEEtran}
\bibliography{example}  
\end{document}